\documentclass[lettersize,journal]{IEEEtran}
\usepackage{amsmath,amsfonts}
\usepackage{algorithm}
\usepackage{algpseudocode}
\usepackage{array}
\usepackage[caption=false,font=normalsize,labelfont=sf,textfont=sf]{subfig}
\usepackage{textcomp}
\usepackage{stfloats}
\usepackage{url}
\usepackage{verbatim}
\usepackage{graphicx}
\usepackage{cite}
\usepackage{tabularx}
\usepackage{hyperref}

\begin{document}

\title{A Novel Bifurcation Method for Observation Perturbation Attacks on Reinforcement Learning Agents: Load Altering Attacks on a Cyber Physical Power System}

\author{Kiernan Broda-Milian, Hanane Dagdougui, Ranwa Al-Mallah
\thanks{The code and data for this work is available on \href{https://github.com/RMC-AIvsAI/A-Novel-Bifurcation-Method-for-Observation-Perturbation-Attacks-on-Reinforcement-Learning-Agents}{Github}.}
\thanks{Supplementary figures and analysis are available in \cite{mythesis}}
}



\maketitle

\begin{abstract}
Components of cyber physical systems, which affect real-world processes, are often exposed to the internet. Replacing conventional control methods with Deep Reinforcement Learning (DRL) in energy systems is an active area of research, as these systems become increasingly complex with the advent of renewable energy sources and the desire to improve their efficiency. Artificial Neural Networks (ANN) are vulnerable to specific perturbations of their inputs or features, called adversarial examples. These perturbations are difficult to detect when properly regularized, but have significant effects on the ANN’s output. Because DRL uses ANN to map optimal actions to observations, they are similarly vulnerable to adversarial examples.
This work proposes a novel attack technique for continuous control using Group Difference Logits loss with a bifurcation layer. By combining aspects of targeted and untargeted attacks, the attack significantly increases the impact compared to an untargeted attack, with drastically smaller distortions than an optimally targeted attack. We demonstrate the impacts of powerful gradient-based attacks in a realistic smart energy environment, show how the impacts change with different DRL agents and training procedures, and use statistical and time-series analysis to evaluate attacks' stealth. The results show that adversarial attacks can have significant impacts on DRL controllers, and constraining an attack's perturbations makes it difficult to detect. However, certain DRL architectures are far more robust, and robust training methods can further reduce the impact. 
\end{abstract}


\section{Introduction}
\IEEEPARstart{C}{yber} Physical Systems (CPS) are networked Operational Technology (OT), which control processes in the real world \cite{Security_2021}. Thus, actions in cyberspace can have physical consequences. There are many precedents for cyber attacks on Critical Infrastructure (CI), using a variety of vectors \cite{Zografopoulos_Hatziargyriou_Konstantinou_2023}. Smart grids are a prominent example of critical CPS. They leverage Distributed Energy Resources (DER), which provide a significant attack surface for bad actors to disrupt a power grid. 

Deep Reinforcement Learning (DRL) is pervasive in CPS control research, including smart energy systems \cite{Cao_Hu_Zhao_Zhang_Zhang_Liu_Chen_Blaabjerg_2020,Perera_Kamalaruban_2021}. These systems require fine grained management to instantaneously match power generation with the load, by controlling several variables and safety systems. As energy systems are a component of CI, operational efficiency and reliability is broadly consequential. DRL addresses the increasing complexity of integrating multiple energy sources, rising demand, and non-linear behaviour. Unlike conventional control methods, DRL agents can learn optimal actions without knowing the environmental dynamics, through interactions with historical data, simulations, or live systems. In contrast, conventional control methods must model the system's dynamics before optimal actions can be determined. RL applications are successfully studied in many power system applications, from the operational control of energy generation and distribution, to the efficient use on the load side, and determining market prices. There is limited research on the robustness of the RL agents in these applications.

Although OT is used for control, it's rife with security concerns given its safety-first focus, which prioritizes uninterrupted communication and continuous operation over security\cite{Security_2021,Zografopoulos_Hatziargyriou_Konstantinou_2023}. OT's exposure to the internet as a part of modern CPS creates additional vulnerabilities. In fact, there are a variety of threats, from ransomware gangs to state-sponsored actors. Thus, protecting these systems and exploring their vulnerabilities is an important area of research. With the numerous applications for DRL, adversarial examples, in concept, provide another exploit to adversaries \cite{Ilahi_Usama_Qadir_Janjua_Al-Fuqaha_Hoang_Niyato_2022}. 

This work addresses the following research questions for a DRL controller in a Demand Response (DR) environment:
\begin{enumerate}
    \item What is the potential impact of adversarial observation perturbations in a white-box setting?
    \item Can \textit{impactful} observation perturbations be statistically indistinguishable from the original observations?
    \item Are there algorithm-agnostic methods for improving DRL robustness in this environment? 
\end{enumerate}

In this work, agents with different architectures were trained in a gym environment \cite{nweye2023citylearn} to reduce energy usage by controlling the charging and discharging of a battery energy storage system. These agents were used to test three types of white-box attack: untargeted, optimally targeted using an adversarial agent, and using a novel method called the bifurcation attack. 
The untargeted attacks had relatively small effects on power consumption, but were successful in changing the agent’s action for nearly every observation. The optimally targeted attack more than tripled the power consumption, but introduced obvious distortions. The novel attack more than doubled the effects of the untargeted attack with similar distortions to the observations. 
Statistical analysis of these adversarial observations indicated that stealthy adversarial attacks are possible and statistically indistinguishable attacks could have a significant impact on their victim. The action space of a DRL agent significantly affects the impact of attacks. In fact, comparing the same DRL algorithm with discrete and continuous action spaces showed that the former reduced the attack's effect by approximately three-fourths.
The black box attack was successful in significantly increasing power usage without access to the agent’s parameters. It involved training a proxy ANN using the agent’s historical observations and actions, and using the proxy’s parameters to enable a simple attack. This attack is more feasible in practice, as this type of historical data would be available to an attacker with access to the agent’s observations. The compromise is that larger distortions are required for the black box attack to achieve a similar effect as a white box attack.

This work makes the following contributions:
\begin{enumerate}
    \item We propose a novel adversarial attack technique for continuous control. By implementing the Grouped Difference Logit (GDL) loss with a bifurcation layer, the adversarial regret for SotA attacks was doubled for continuous control agent in CityLearn.  The bifurcation method can be used for continuous and discrete outputs, even for attacks which only support Artificial Neural Networks (ANN) with multiple outputs. This technique was validated using identical attacks without the bifurcation layer for both classification/discrete actions and regression/continuous actions. To the best of the author's knowledge, no other work has used this or a similar technique. 
    \item With a carefully selected budget, distributions of adversarial observations, which are not significantly different from the originals, cause a significant adversarial regret. Conversely, adversarial observations form a distinct distribution when considering the absolute difference between subsequent observations in a time-series. To the best of the author's knowledge, no other work has studied the detection of adversarial observations (adversarial examples) in a CPS gym which uses real world data. 
    \item We demonstrate that the bifurcation method improves the black-box Snooping attack \cite{Inkawhich_Chen_Li_2020}, and that ATLA \cite{Zhang_Chen_Boning_Hsieh_2021} is an effective defence against this new attack.
\end{enumerate}

The remainder of this paper is organized as follows. Section \ref{sec:related work} reviews related works on adversarial machine learning and RL and security for smart grids. Section \ref{sec:threat} discusses the threat model. Section \ref{sec:methodology} describes the methodology. Section \ref{sec:results} presents the results. Section \ref{sec:conclusion} is the conclusion.

\section{Related Work}
\label{sec:related work}

Proximal Policy Optimisation (PPO) \cite{schulman2017proximal} is an on-policy policy-gradient algorithm which can operate with continuous and discrete action spaces. The PPO is unique for the surrogate loss function used to train its policy, which prevents large changes during each update. An on-policy algorithm like PPO learns from experiences of actions sampled from its current policy alone, while an off-policy algorithm like Soft Actor-Critic (SAC) learns from the stored experiences of previous policies \cite{Haarnoja_Zhou_Abbeel_Levine_2018}. The SAC is unique for its combination of off-policy actor-critic training, with a stochastic actor with an entropy maximisation objective to encourage exploration. Although the SAC outperforms the PPO in various continuous control tasks, it is limited to continuous action spaces.

The surveys \cite{Cao_Hu_Zhao_Zhang_Zhang_Liu_Chen_Blaabjerg_2020,Perera_Kamalaruban_2021}, provide comprehensive reviews of recent research on RL applications in energy system management. They show the potential of RL in future energy systems.  Due to uncertainties in renewable energy generation and demand, RL outperforms other techniques by addressing three main issues \cite{Perera_Kamalaruban_2021}:
\begin{enumerate}
        \item Nonlinear systems and nonconvex optimisation functions in modelling.  RL algorithms are able to operate without models and can learn to approximate their environment.
        \item Nonexistence of physical models, using model-free learning as above. This is a common issue with building energy management systems which typically lack practical models.
        \item Operating in large state and action spaces is less challenging for RL, for example, using weighted function approximations can result in an ANN learning a set of weights which is much smaller than the set of the state or action spaces.
\end{enumerate}
The work of \cite{Li_Yan_Xie_Sang_Yuan_2019} provides an overview of the cyber attack surface presented by cyber physical power systems and categorizes potential attacks. Discussions on the types of vulnerable infrastructure that could enable Load Alternating Attacks (LAA) are of particular relevance to this work. Further information of different types and effects of LAA is found in \cite{Amini_Pasqualetti_Mohsenian-Rad_2018}. LAAs enabled by compromised Internet-of-things devices were simulated using the architecture of the Polish power grid in \cite{Soltan_Mittal_Poor_2018}. Their results show that cascading power grid failures could be caused by an attacker controlling less than 5\% of total load, if the attack is properly timed. These works are complemented by \cite{Zografopoulos_Hatziargyriou_Konstantinou_2023}, which provides a review of present vulnerabilities of DERs, including attack vectors and cyber attacks on CI, that could enable LAAs. It identifies several unsecure communication protocols widely used in North American power grids, which could allow an attacker to modify transmitted data. It also highlights exploits in Internet-of-things devices that have been used in cyber attacks and characteristics that make them vulnerable to future attacks.
 
The survey in \cite{Standen_Kim_Szabo_2023} describes several frameworks for understanding adversarial DRL. These include: Markov decision processes for modelling an adversary, adversarial goals, and attack vectors. Following this framework, this work uses observation perturbations. Several techniques have been demonstrated to optimize the size and number of perturbations used in adversarial observation perturbation attacks \cite{Ilahi_Usama_Qadir_Janjua_Al-Fuqaha_Hoang_Niyato_2022,Behzadan_Munir_2017}. The techniques also try to reduce the time required to generate the perturbations \cite{Kos_Song_2017,Lin_Hong_Liao_Shih_Liu_Sun_2019,Hussenot_Geist_Pietquin_2020,Mo_Tang_Li_Yuan_2022}. An optimally targeted adversarial attack on a DRL agent was demonstrated in \cite{Behzadan_Munir_2017}. A victim agent was induced to follow an adversarial policy using targeted observation perturbations. The targets are chosen by a DRL agent which is trained in the victim's environment with negative rewards. Thus, the adversarial policy selects actions to minimize the reward.
In this work the methodology for statistical detection proposed in \cite{Grosse_Manoharan_Papernot_Backes_McDaniel_2017} is used in this work for agent-agnostic detection. It relies on the Maximum Mean Discrepancy (MMD) test proposed in \cite{Gretton_Borgwardt_Rasch_Schölkopf_Smola_2012}. Because MMD detects weak attacks but is bypassed by stronger techniques \cite{Carlini_Wagner_2017}, it can validate if an attack is trivially detectable. 
There are few examples of research on adversarial observation perturbations in DRL controlled CPPS, and they do not test if their perturbations are trivially detected. Adversarial attacks on DRL agents were carried out in operational control gym environments of the power grid with a black-box threat model in \cite{Chen_Arnold_Shi_Peisert_2021}. Alternating Training with Learned Adversary (ATLA) was used as a defence against a Learned Adversary (LA) attacks on a MARLISA algorithm controlling multiple buildings in a CityLearn demand response environment \cite{Zeng_Qiu_Sun_2022}.

\section{Threat Model}
\label{sec:threat}
Vulnerabilities in Artificial Neural Networks (ANN) are similar to unpatched exploits in software. This research is a proof-of-concept for exploiting these vulnerabilities in DRL specific techniques to conduct LAAs.  The primary threat model for the attack is a DRL agent that is a white-box, meaning that the attacker can read its parameters and is able to modify sensor readings sent to the agent. The white-box model demonstrates the strongest attacks available, establishing the risk ceiling. Knowing the ceiling informs smart energy designers about the level of risk they are accepting by using vulnerable agents. Furthermore, analyzing such strong attacks can suggest mitigation techniques and quantifying risk is an important step towards mitigation. 
A well resourced attacker could obtain the victim agent parameters without the privileges required to compromise the entire DR system. Possible methods would be compromising a back-up server, compromising an employee to exfiltrate the agent, or obtaining a low level of access on the host or server running the controller DRL which is able to read the agent's parameters. Enabling a black-box snooping attack \cite{Inkawhich_Chen_Li_2020} is even simpler, the attacker only needs historical data to train a proxy ANN. There are precedents for poor security in CI, and security DR infrastructure would be even more fraught when it is maintained by the user, who may lack the resources of a utility to secure it. 

CPPSs are based on distributed sensors that transmit measurements to a controller. Injecting observation perturbations can be accomplished using multiple vectors depending on the exact application. Both the Sunspec Modbus and IEEE 1815 (DNP3) protocols are vulnerable to intermediary attacks, allowing an attacker with a presence on the LAN to capture and modify data in transit \cite{de_Carvalho_Saleem_2019}. Many smart consumer energy devices have known software vulnerabilities \cite{Soltan_Mittal_Poor_2018}, some of which have already been exploited to form bot nets \cite{Zografopoulos_Hatziargyriou_Konstantinou_2023}. The diverse manufacturing stages of smart devices also make them vulnerable to supply chain attacks, attacks might physically access sensors or means of transmission in unsecure public or isolated locations. These vulnerabilities can give attackers a foothold to modify traffic, or compromise sensor firmware to add perturbations. Given the vulnerabilities and precedents discussed above, an attacker could gain the access to conduct a LAA using adversarial observations. This level of access is less than what would be needed to compromise the DRL controller itself. Furthermore, adversarial observations have the potential to conceal the cause of the attack. For this research, the attacker is assumed to have write access to the observations of a DRL agent, and in some cases access to the agent ANN's parameters.

\section{Methodology}
\label{sec:methodology}
In this section, we will discuss the tools required for the experiments carried out in this study. We will then present different white-box attacks conducted against different agents. We will detail the proposed bifurcation and target group approach devised to increase the impact of untargeted attacks without significantly increasing the perturbation magnitude. We will validate that these attacks are not trivially detected, and test two methods of increasing the victim agent's robustness. Finally, we will demonstrate black box attacks.

\subsection{Gym Environment}

CityLearn has been selected as it is a well-documented and actively maintained smart energy DR gym, following OpenAI Gym standards \cite{Vazquez-Canteli_Dey_Henze_Nagy_2020}. Its environments feature districts with customisable quantities and types of buildings whose energy demands are set using historical data. In the 2022 challenge environment, one or more agents learn to charge and discharge building electrical storage to reduce and smooth demand from the electrical grid. It includes up to five residential buildings each with their own battery. The observations are generated using real-world data from a California study on residential photovoltaic integration \cite{nweye2023citylearn,Morell_2017}. The environment's 31 features include (* indicates corresponding 6, 12 and 24 hour prediction features):
\begin{enumerate}
    \item Outdoor dry bulb temperature*;
    \item Outdoor relative humidity*;
    \item Diffuse and direct solar irradiance*;
    \item Carbon intensity;
    \item Non-shiftable load;
    \item Solar generation;
    \item Electrical storage State of Charge (SoC);
    \item Net electricity consumption; and
    \item electricity pricing*.
\end{enumerate}
Because the victim's observations are generated from recorded data, the attack's observation perturbations will be generated is a realistic setting. This realism enables testing the hypothesis that adversarial examples would be a difficult payload to detect during a real attack. Due to the restrictions of SotA white-box adversarial attacks, the victim agent controls the battery's State of Charge of a single building so that the action space remains one-dimensional. This frames the RL task as a classification problem.

 The Key Performance Indicators (KPI) are the results of a cost function over the course of an episode, normalized by the system's performance without a DRL controller. Lower KPIs are better, and they indicate the proportion of improvement caused by the DRL controller. The most important KPI for assessing impact of an attack on grid stability is electricity consumption, as blackouts result when power echange with the grid exceeds grid capacity. This would require a large-scale attack, either on many victims or a high-consumption victim like an industrial park. Peak consumption is the second most important KPI, as a significant peak during peak grid demand could also cause a blackout.  Ramping is an important KPI as the grid must instantaneously match demand, so large fluctuations therein are challenging for the grid operator. However, if the ramping is not accompanied by significant energy consumption or high peaks, it is unlikely to affect grid stability.

\subsection{Untargeted and Targeted Attacks}

We start by conducting SotA untargeted gradient-based attacks on the agents trained in the previous section. The attack's success is measured in terms of its Adversarial Success Rate (ASR), and adversarial regret for the electricity consumption, ramping, and daily peak KPIs. The attacks used are the Maximum Confidence Auto-Conjugate Gradient (ACG) \cite{Yamamura_Sato_Tateiwa_Hata_Mitsutake_Oe_Ishikura_Fujisawa_2022} and the Minimum-Norm Brendle and Bethge (BB) \cite{Brendel_Rauber_Kümmerer_Ustyuzhaninov_Bethge_2019}, both implemented in the Adversarial Robustness Toolbox \cite{art2018}. All attacks are carried out using $L_\infty$ regularisation, where both attacks demonstrate the best performance. For a maximum-confidence attack, $L_\infty$ regularisation clips the perturbation for each feature in the range $[-\epsilon, \epsilon]$ and is the least restrictive regularization method. As these attacks are designed for classifiers, a discrete action space PPO was trained as the victim.
The maximum-confidence ACG attack is enhanced using multiple restarts with different $\epsilon$, to minimize the size of the adversarial perturbation. An $\epsilon$ is selected from a list of candidates using a binary search. This technique is referred to as the \textit{dynamic distortion}.

Untargeted attacks force the victim agent to take any suboptimal action. In a control environment like CityLearn, a small change to the input action may only result in a small change to the energy stored or discharged. Suboptimal actions do not necessarily cause significant increases in energy consumption. By inducing the victim to take the worst possible action at any timestep, adversarial regret can increase significantly compared to an untargeted attack. To select the optimal worst action, a DRL agent was trained to choose the action which minimized rewards in CityLearn. This represents a targeted attack. A PPO with hyperparameters identical to the victim was trained with a negative of the victim's reward, in a reward-based attack. In this case the adversarial policy maximizes power usage, rather than minimizes it. For each observation, the adversarial policy selects the optimally worst action, and this is the target for the adversarial attack. This is called \textit{policy induction}. Thus a successful attack induces the victim into taking the optimally worst action, which increases the adversarial regret compared to an untargeted attack. 

\subsection{Bifurcation and Target Group}
\begin{figure}
    \centering
    \includegraphics[width=1\linewidth]{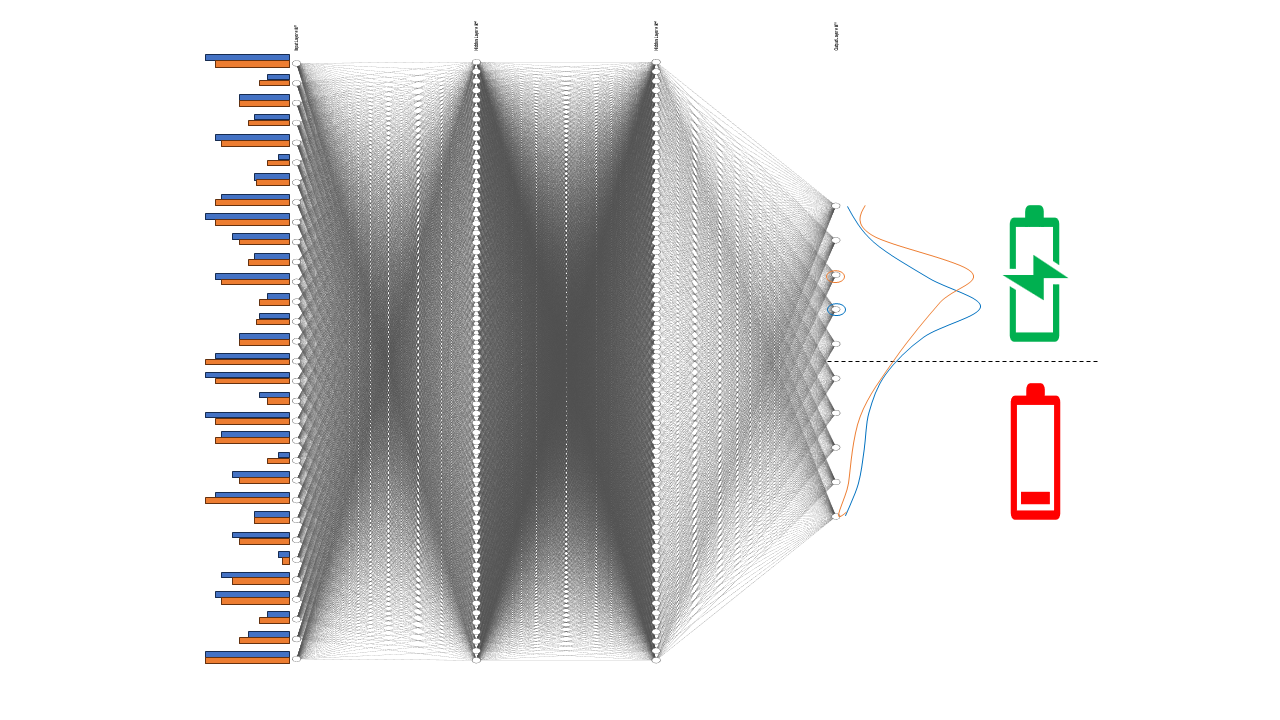}
    \caption{Example of an adversarial attack on a discrete actor network trained in CityLearn. The original observations and actions are represented by elements in blue, and adversarial in orange. The bars on the left represent features in an observation, and the curves on the right represent the value of each logit (which correspond to actions). The upper five logits represent different levels of charge actions, and the bottom five discharge. In this example, small changes to the original observation result in an adversarial action different from the original. But, the result is only slightly more charging than is optimal, so the impact on power consumption is limited.}
    \label{fig:atk-ex}
\end{figure}

Figure \ref{fig:atk-ex} shows how untargeted attacks don't necessarily lead to significant changes in the victim agent's behaviour. This can be addressed with optimally targeted attacks, but these have two principle disadvantages: training an adversarial policy, and targeted attacks can require significantly larger distortions. Grouped targeting is devised to increase the impact of untargeted attacks without significantly increasing the perturbation magnitude. 

\subsubsection{Grouped Difference Logit Loss}
Consider this simple function, Difference Logit (DL) loss, which is similar to the loss used in \cite{Brendel_Rauber_Kümmerer_Ustyuzhaninov_Bethge_2019}:
\begin{equation}
\label{eq:DL}
        DL(x,y) = z_y - \max_{i\neq y} z_i
\end{equation}
An attacker can use this loss function to craft an adversarial sample corresponding to any label excluding the original ($y$).  However, this is not helpful for the problem at hand, as the untargeted attacks demonstrated that small changes to the agent's actions do not significantly affect its performance. But as above, nor is a targeted attack helpful, as there are multiple acceptable adversarial labels or actions. Consider the group $G$ which is a subset of the agent's actions, a loss function that maximised any label in $G$ over the original would be applicable to the problem at hand.  This Grouped DL (GDL) loss can be written as:
\begin{equation}
        GDL(x,y) = z_y - \max_{i\in G} z_i
        \label{eq:GDL}
\end{equation}
Like the target, the group $G$ changes over the duration of an attack on a DRL agent. Specifically, when the agent chooses one of the charge actions, all discharge attacks are added to the target group.  This approach is ideal for control tasks where the distance of an adversarial action from the optimal action is proportional to the adversarial regret it causes. Consider a steering input for a vehicle; the further the input is from optimal, the further it will stray off course. In classification tasks generally, any label but the correct one has an equal adversarial regret in terms of accuracy. However, there are circumstances where certain groups of misclassifications result in a larger adversarial regret than others. Consider an image classifier used by an autonomous vehicle: causing a Sedan to be misclassified as a truck or bus might have some effect on the vehicle's behaviour, but misclassifying it as any non-vehicle could have an even greater effect.
\subsubsection{Bifurcation Layer}
\begin{figure}
    \centering
    \includegraphics[width=1\linewidth]{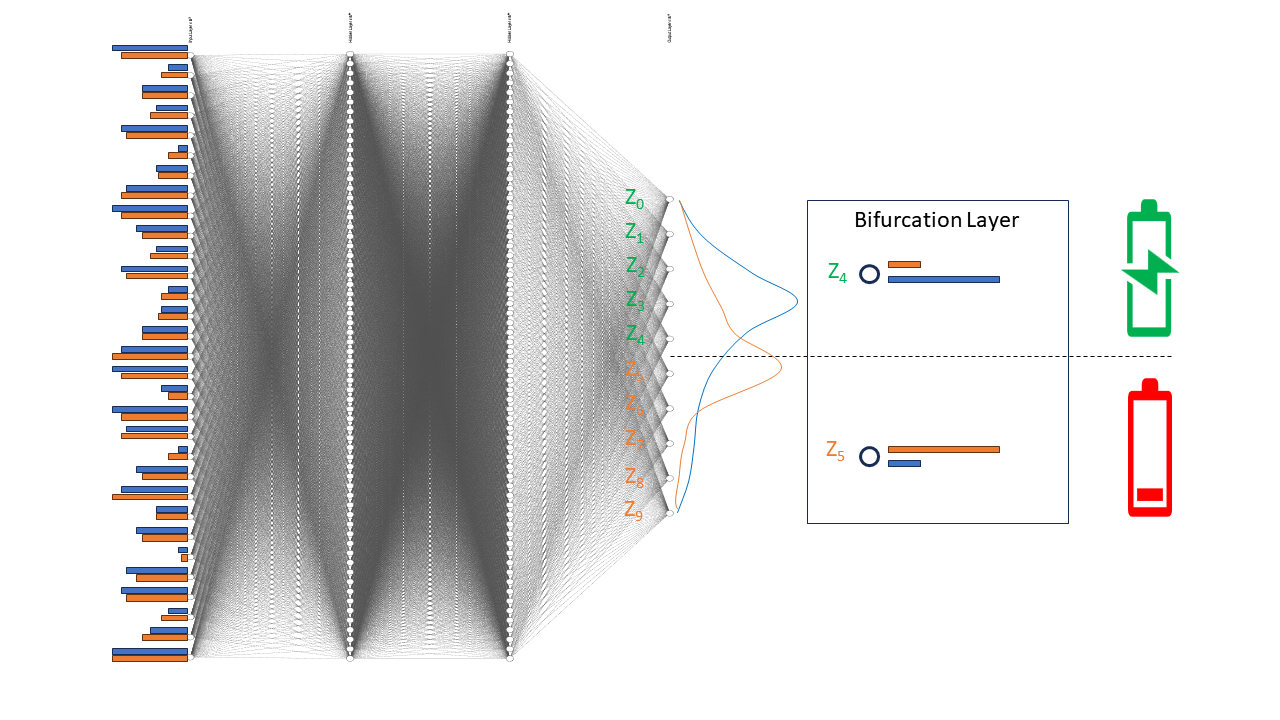}
    \caption{Example of an adversarial attack on a discrete actor network trained in CityLearn, using the bifurcation method. The original observations and actions are represented by elements in blue, and adversarial in orange. The bars on the left represent features in an observation, and the curves on the right represent the value of each logit (which correspond to actions). The upper five logits represent different levels of charge actions, and the bottom five discharge. The output of the bifurcation layer is the maximum logit value for each of these groups of logits. In this example, small changes to the original observation result in a discharge action instead of the original charge action. Inducing the victim agent to reverse its (dis)charge decisions increases electricity consumption.}
    \label{fig:bi-atk-ex}
\end{figure}

The actor network for a discrete agent in CityLearn will have symmetrically graduated actions for various levels of (dis)charge. These can be combined into target groups for GDL loss by modifying the network used by the adversarial attack to return only a pair of logits, corresponding to the maximal charge and discharge logits. This binary maximum layer makes the DL loss equivalent to the GDL loss. We call this layer the bifurcation layer, which is described in Figure \ref{fig:bi-atk-ex}. Unlike GDL, DL is easily implemented in ART using PyTorch. By implementing GDL through the input network, the method is practical for most adversarial frameworks and methods, rather than restricting it to a single library or attack. 
\subsubsection{Continuous Action Spaces}
Using the bifurcation method an adversary changes the victim ANN's outputs to suit their objective. Above, logits are grouped to surrogate targets, but the same can be done to add logits to a regressor with only a single output. The majority of adversarial attacks are designed for classifiers, for which one logit corresponds to a label and the largest logit value is the ANN's prediction. A regressor has a single output with a continuous range of values. These correspond to discrete and continuous action spaces in DRL, and continuous action spaces are typically used for control tasks such as the DR simulated in CityLearn. This limitation in adversarial example research significantly limits their application to DRL controllers. However, a logit layer can be added to a regressor as glue logic between the ANN and attack algorithm.  
The simplest bifurcation layer tested consists of two outputs, one is the same as the regressor's original output, and the other its negative. Instead of returning a single prediction value, the network is induced to return a tuple of two values.  The output $y$ becomes the vector $(y, -y)$.  Thus the greater of the two logits is treated as the ANN's prediction. The loss gradient will be used to reduce the original logit while increasing the negative logit. It does not matter if the greatest logit changes during the attack, as the attacker's goal is to find inputs which significantly changes the regressor's or agent's prediction. Other bifurcation layer configurations were explored, using various exponential and linear functions, but none were found to outperform this method in preliminary experiments.

Minimum-norm attacks are inappropriate for this task because they rely on a classifier's decision boundary, the boundary is crossed when the logit with the largest value changes. There is no analogous concept for a regressor with a single output.  Maximum-confidence attacks simply maximize an arbitrary loss function, so they are easily adapted for regression. Instead of maximising the value of a logit other than the original, they maximise the difference between the original and adversarial outputs within their budget. The attacker can change the adversarial budget to balance adversarial regret and stealth.

To compare the bifurcation method to a direct attack on an agent with a continuous action space, an attack is required that is compatible both with the continuous output of the agent and the bifurcated output. Publicly available implementations of ACG can only do the latter, as it amounts to a classification. To this end, a simple Projected Gradient Descent (PGD) \cite {Madry_Makelov_Schmidt_Tsipras_Vladu_2018} attack for arbitrary loss functions is implemented, which is shown in Algorithm \ref{alg:myPGD}. This PGD implementation allows attacks using loss functions such as Mean Squared Error (MSE), Mean Absolute Error (MAE), and Huber loss, which are compatible with the regression networks used by continuous action agents. This attack is used to compare the adversarial regrets for the SAC agent with and without the bifurcation layer, and between the SAC and discrete PPO. In addition to adversarial regret, there are several metrics which are useful in comparing the effects of these attacks:
\begin{enumerate}
    \item Because there is no discrete boundary between the decisions of a regressor or continuous agent, MAE will be used instead. The discrete agent's actions are mapped onto a continuous action space, so the MAE can be calculated for both. MAE is linear and does not alter the unit of the inputs, so it can be intuitively understood as the \textit{distance} between the original and adversarial actions.
    \item The (dis)charge reversal is proportional to the adversarial regret, and the goal of the bifurcation method. The early untargeted attacks had a limited effect on this metric, and similarly limited adversarial regrets. By comparing the signs of the original and adversarial actions, the proportion of timesteps where the victim agent's decision is reversed can be counted. 
\end{enumerate}

\begin{algorithm}
    \caption{PGD with Decaying Stepsize \cite{Madry}.}
    \label{alg:myPGD}
\begin{algorithmic}
    \State \textbf{Inputs:} ANN $\theta$, sample $x$, ANN prediction $y$, loss function $L$, $\epsilon$, stepsize $\eta$, iterations $N_{iter}$, decay rate $\alpha$, number of decays $N_\alpha$
    \State $\delta^0 \gets 0$
    \State $k_\alpha \gets \lfloor\frac{N_{iter}}{N_\alpha}\rfloor$
    \For{$k=1 \text{ to } N_{iter}$ }
        \State $s^k \gets sign(\nabla L(x+\delta^{(k-1)},y,\theta))$
        \State $\delta^k \gets P_\epsilon(\delta^{(k-1)} + \eta s^k )$
        \If{$k \mod k_\alpha == 0$}
            \State $\eta \gets \eta\alpha$
        \EndIf
    \EndFor
    \State \textbf{return} $\delta^k$
\end{algorithmic}
\end{algorithm}

\subsection{Detection}
The purpose of detection in this work is to determine if the set of adversarial observations during one episode of an attack are plausible in aggregate.  Determining if an individual sample or observation is adversarial is outside the scope of this work.  Detection has dual purposes, both in defending against adversarial observation and in proving whether an attack is feasible. While the literature reviewed in this work suggests several novel methods for attacking DRL agents, none tested if their attacks were in fact stealthy. The aim of this section is to determine if gradient-based adversarial attacks can generate plausible adversarial observations.  

To do so, we take a two pronged approach to ensure the plausibility of adversarial observations. First, the Maximum Mean Discrepancy (MMD) Gaussian kernel detector \cite{Viehmann_Antiga_Cortinovis_Lozza_2024} is used to evaluate if the set of adversarial observations is plausible, following the methodology of \cite{Grosse_Manoharan_Papernot_Backes_McDaniel_2017}. Then, aggregated time-series analysis is used to analyse variations in features over time. In this way both the construction of individual samples, and their relationship with the preceding sample is analyzed. We use agent-agnostic statistical techniques to identify outlying features of adversarial observations, and statistical testing to determine if the original and adversarial observations form distinct distributions. 

This analysis is enabled by the real-world measurements used to generate CityLearn's observations, because detection is tested in a realistic control setting. This will determine if the adversarial attacks developed are feasible. An attack is only considered feasible when the adversarial observations it generates are plausible and the attack causes significant adversarial regret. 
Unlike the image classification setting where adversarial attacks were first developed and are typically studied, observations in a DRL environment have the following characteristics which enable detection:
\begin{enumerate}
    \item Correlated features: the weather conditions are dependant on the time or day and are loosely periodic. e.g., temperatures and solar irradiance peak at mid day and drop at night. Adversarial observations which fail to follow these patterns could be detected simply by plotting them.
    \item Observations in a DRL environment are a time series, which means there is a relationship between subsequent observations. So, adversarial observations must not only be individually plausible, but also be plausible given the previous observation. e.g., weather features for a particular climate will vary within a particular range over time, so larger temperature variations between measurements could indicate false data injection. The mean rate and range of inter-observation changes for clean observations is compared to those for adversarial observations, to test if they can be separated by some threshold value. 
\end{enumerate}
Based on this analysis of adversarial observations, the attack's adversarial budget will be reduced until the observations it produces appear plausible. Attacks in this work use an $L_\infty$ boundary, which restricts the maximum perturbation for individual features. This size of $\epsilon$ directly affects the amount of distortion between original and adversarial observations, so reducing it can improve an attack's stealth.

MMD is a statistical method of identifying if two sets of samples are drawn from the same distribution. The MMD test as used in \cite{Grosse_Manoharan_Papernot_Backes_McDaniel_2017} will determine if the distribution of adversarial observations is plausible compared to the originals.  Because it is made for sets rather than individual samples, it does not detect individual adversarial samples. However, it can be used as a metric to assess the significance of the adversary's distortions by comparing the clean and adversary observations generated during an episode. This represents the best case scenario for detecting the adversary's distortions, because the "correct answer" is used in the form of original observations. This would not be the case in a production system, as present observations can only be compared to historical data, and changing patterns in weather and electricity demand can also cause drift which increases MMD.

The purpose of this test is to suggest a threshold of distortion ($\epsilon$) which is likely to be detectable with agent agnostic statistical methods. If an attack exhibits a low MMD in this ideal setting, then it is unlikely to be detected statistically. In \cite{Carlini_Wagner_2017}, the authors demonstrate that MMD will not detect strong minimum norm attacks, while \cite{Grosse_Manoharan_Papernot_Backes_McDaniel_2017} showed it is effective for weak attacks like the Fast Gradient Method (FGM) \cite{Goodfellow_Shlens_Szegedy_2015}. These results make MMD valuable for determining whether SotA attacks are indeed plausible, because the test can be evaded by a strong adversary. 

Using the same methodology as \cite{Grosse_Manoharan_Papernot_Backes_McDaniel_2017}, the MMD is calculated using 10 000 bootstraps with a Gaussian kernel. The test provides the MMD value which quantifies the difference between two distributions, and a probability that both distributions are the same. Due to the correlated nature of CityLearn's time series observations, finding a clean MMD threshold is more complicated than simply randomly sampling a clean distribution as done in \cite{Grosse_Manoharan_Papernot_Backes_McDaniel_2017}. Through experimentation, the observations of a clean episode are divided into two representative samples, and the results of this test will be used as a baseline for clean data. If the MMD test results between the original and adversarial observations in an episode are outside the baseline range, the adversarial observations are considered implausible.

\subsection{Robust Training}
Two methods of agent robustness are explored: Alternating Training with Learned Adversary (ATLA) \cite{Zhang_Chen_Boning_Hsieh_2021}, and binning continuous action spaces.

ATLA is a training method that teaches an agent a robust policy ($\pi$). It was chosen for testing this work because it can be applied to any DRL algorithm, potentially making it an accessible method to defend against adversarial attacks.  The literature review of this work found no evaluation of resistance to adversarial attacks targeting a DRL agent's function approximator, so that is the goal of this experiment. 

The environments for the adversary and victim use identical CityLearn parameters but have different action spaces and rewards. The adversary receives a clean observation from CityLearn and its action is adding a perturbation, then the victim's static policy in the adversary's environment selects an action based on the adversarial observation. Conventionally, the action space of the adversary is constrained by a function $A_{adv} \in B(s)$, where $s$ is the current state. To avoid violating the Markov property and be consistent with the $L_\infty$ constraint from previous attacks, $B(s)$ is a static boundary for each feature. This boundary will be generated using typical variations between observations during a clean episode. $B(s)$ is a training hyperparameter and must be selected such that the adversary is capable of a significant adversarial regret without preventing the agent from learning.

In addition to robust training with ATLA, the discrepancy in robustness between the SAC and the discrete PPO will be investigated. Because there are two variables between these agents, the learning algorithm and the action space, they will be isolated by evaluating the robustness of a continuous action space PPO. The adversarial regret of all three agents for the stealthy attack defined in the detection section will be compared.

\subsection{Black Box Attack}
This section tests the impact of the black-box snooping attack on the same agents used to conduct white-box attacks, to show that adversarial attacks can decrease the performance of a DRL controller without access to its ANN parameters. 
The snooping attack \cite{Inkawhich_Chen_Li_2020} allows an attacker to leverage adversarial attacks with \textit{no knowledge} of the victim agent's algorithm or architecture, unlike previous white box attacks which assume the attacker was able to copy the ANN parameters. This is also different from training a surrogate ANN, which is done using the victim agent's environment. Instead, the attacker is only able to observe the victim agent acting in its environment, which is a requisite level of access for conducting an observation perturbation attack. If the attacker can change the victim's observations, it is probable that they can also observe them.
Like the snooping attack, a replay attack can also be conducted with access to historical data and the ability to modify the victim's observations. There are several advantages to the snooping attack:
\begin{enumerate}
    \item Adversarial examples can be crafted when some features cannot be changed. This is an advantage when the attacker can perturb most but not all features e.g. is able to change sensor readings but not the internal time on the controller.
    \item The attacker must collect observations which correspond to the desired actions to conduct a replay attack. This is an issue when an agent rarely takes the desired action or only takes it in a state very different from the present. The snooping attack can craft adversarial observations on the fly.
    \item Using the snooping attack, perturbations can be constrained to suit the target environment. Despite being statistically distinguishable from normal samples, FGM perturbations can still be relatively small and could feature smaller inter-sample changes compared to disjointed replay samples. 
\end{enumerate}
Using the State-Action threat model, an ANN proxy imitator was trained, which enabled FGM attacks. Like previous white-box attacks, the bifurcation method is compatible with a proxy and the FGM attack. A bifurcation layer was appended to the trained proxy, in the same fashion it was added to a trained agent's policy network. The proxy used no prior architectural knowledge of the agent, instead hyperparameters were chosen during an Optuna \cite{Akiba_Sano_Yanase_Ohta_Koyama_2019} trial which maximized the mean time series split validation over one episode of data. 

\section{Results}
\label{sec:results}
\subsection{White Box Attacks}
A PPO agent with 20 discrete actions and a SAC were trained for 500 episodes. We refer to the former as the discrete PPO. Table \ref{tab:PPOd_atk_kpi} shows the performance of a discrete PPO agent under various conditions. Despite the untargeted attacks changing $\sim 99\%$ of the victim agent's actions,  there is not a large increase in the KPIs from these attacks. The untargeted adversarial attacks are successful, but this does not significantly affect the building's power consumption. The optimally targeted attack is extremely effective, at least tripling every KPI. However, this attack results in perturbations that are an order of magnitude larger than the untargeted attacks. For example, temperature readings are changed by several degrees at each observation. The optimally targeted attack is obvious from the adversarial observations.

\begin{table}
    \centering
\caption{Adversarial attack KPIs for Discrete PPO. KPI are normalized measurements of cost functions in CityLearn.}
 \label{tab:PPOd_atk_kpi}
    \begin{tabularx}{\columnwidth}{|X|X|X|X|} \hline 
         \multicolumn{4}{|>{\centering\arraybackslash}p{\dimexpr\columnwidth-5\tabcolsep-3\arrayrulewidth\relax}|}{Adversarial Attack KPIs for Discrete PPO}\\ \hline 
         KPI &  Electricity Consumption &  Daily Peaks &  Ramping \\ \hline 
         Clean&  0.88&  0.89&  1.09\\ \hline 
         Untargeted Dynamic Distortion ACG&  0.91&  0.96&  1.27\\ \hline 
         Untargeted BB&  0.93&  1.01&  1.33\\ \hline 
 Optimally Targeted BB& 3.27& 2.93&17.46\\ \hline
    \end{tabularx}

\end{table}

Table \ref{tab:bi_atk_kpi} shows the results of the PGD attack described in Algorithm \ref{alg:myPGD} with the parameters in Table \ref{tab:PGD_parameters}. Because open source implementations of ACG are not compatible with regressors, a PGD attack was used for comparison. Interestingly, this attack outperforms the ACG with bifurcation. For both continuous and discrete action spaces, PGD using the bifurcation method has a significantly greater impact than PGD alone. For both agents, the bifurcation method results in a difference adversarial (dis)charge from the original. For example, the agent is induced to charge the battery when it originally intended to discharge. The bifurcation method is effective in reversing the victim's discharge action, and this leads to larger adversarial regrets. The discrete PPO was more robust than the SAC. The rate of (dis)charge reversal for the direct attack on the SAC was similar to the discrete PPO's value for the bifurcated attack. 
\begin{table}
    \centering
\caption{Parameters for the PGD attack used in this section ($\epsilon$ varies in the following section).}
\label{tab:PGD_parameters}
    \begin{tabular}{|c|c|c|c|c|} \hline 
         $\epsilon$&  Initial Stepsize&  Iterations&  Stepsize Decays& Decay Rate \\ \hline 
         0.05&  0.01&  100&  4&  0.5\\ \hline
    \end{tabular}

\end{table}

\begin{table}
    \centering
\caption{KPIs for PGD Attacks with $\epsilon=0.05$. This table shows the increased impact of attacks using the bifurcation method and the robustness of the discrete PPO compared to the SAC. The (dis)charge reversal represents the proportion of timesteps where the adversarial (dis)charge action did not match the original.}
\label{tab:bi_atk_kpi}
    \begin{tabular}{|c|c|c|c|c|} \hline 
         \multicolumn{5}{|c|}{KPIs for PGD Attacks with $\epsilon=0.05$}\\ \hline 
         &  \multicolumn{2}{|c|}{SAC}&  \multicolumn{2}{|c|}{Discrete PPO}\\ \hline 
         Metric&  Direct&  Bifurcated&  Direct& Bifurcated\\ \hline 
         Electricy Consumption 
&  1.02&  1.24&  0.89& 0.96\\ \hline 
         Daily Peaks 
&  1.11&  1.93&  0.92& 1.14\\ \hline
 Ramping & 1.12& 2.06& 1.16&1.52\\\hline
 (Dis)Charge reversal& 27\%& 95.7\%& 4.2\%&26.1\%\\\hline
    \end{tabular}

\end{table}

\subsection{Detection}
This section presents the results of the time series feature variation analysis and MMD Gaussian Kernel Detector.
\subsubsection{Time Series Feature Variation Analysis}

Through reducing the adversarial budget and observing the adversarial regret, the parameters for a stealthy attack were selected. The adversarial budget was reduced to decrease the relative proportion of the adversarial mean absolute feature variation compared to the original. The absolute feature variations were analyzed because the changes in mean feature variations and values are negligible with this attack and unlikely to be detected. The absolute feature variation is the absolute difference between a feature and the values from the previous observation. Negative and positive changes can negate each other's effect on the mean, so absolute values are used to measure the magnitude of the variation.

An attack is used to assess the detectability of adversarial observations. An attack on the SAC was chosen, as it quickly became apparent that the adversarial regret dropped significantly faster for the discrete PPO for similar budgets. The PGD attack was used as it allows an  $\epsilon$  to be assigned individually for each feature. The $\epsilon$ was reduced for the solar generation and net electricity consumption features because they were not normalised the same way as other features. This resulted in perturbations many times larger than the feature's value when a uniform $\epsilon$ was used. Thus, the stealthy PGD attack had the following characteristics:
\begin{itemize}
    \item PGD was used with the bifurcation method on a SAC victim agent
    \item $\epsilon=0.03$ for all features except:
    \begin{itemize}
        \item Solar generation and all temporal features had $\epsilon=0$, 
        \item Net electricity consumption has $\epsilon = 4.8\times10^{-4}$, which is the product of its spread and the $\epsilon$ for all other features.
    \end{itemize}
    \item  Each feature was constrained between [0,1]
\end{itemize}
This  bifurcation method improves the adversarial regret by 50\% compared to a direct PGD attack with the same parameters. Constraining the adversarial features to [0,1] reduced adversarial regret by more than a third, which was greater than the reduction for masking or scaling the solar generation and net electricity consumption features. However, the lack of such constraints makes adversarial observations obvious, particularly when a value like solar irradiance or power generation becomes negative.

Despite the adversarial observations being very close to the originals used to craft them, the difference between two adversarial observations is greater than the originals. The stealthy PGD attack increases the adversarial regret for power consumption compared to a direct attack by 50\%, but is limited by the adversarial budget as the regret scale with $\epsilon$. Further decreasing the budget would almost eliminate the advantages from the bifurcation method. Because the budget cannot be reduced further, this section will analyse if the stealthy PGD attack is easily identified. 

For the features with the highest mean absolute inter-observation variations, the original and adversarial distributions of values overlap, with fewer than 10 outliers. 97.6\% of absolute inter-adversarial observation difference are within the original range, and only 15.3\% have more than one outlying feature. This indicates that the distributions are difficult to separate, though it may be possible in higher dimensional space using machine learning with each features' absolute variation. Variations of the sizes introduced by the stealthy PGD attack would  be non-trivial to detect, but the changes in observations are distributed differently from the originals. Further analysis of the difference between the original and adversarial distributions is presented in the following section. 

 \subsubsection{MMD Gaussian Kernel Detector}
The baseline for normal MMD was determined using the data from a clean evaluation episode. Due to the seasonality in CityLearn observations, where both weather and usage feature vary by season, MMD will indicate that different fractions of observations from a clean episode are not drawn from the same distribution, e.g., winter observations follow a different distribution than summer.  Because the MMD is a comparison of two distributions, these clean data needed to be split into two representative samples. Multiple episodes could not be used because the majority of CityLearn's features are deterministic, and the others are influenced by the deterministic policy of the agent, meaning there is no significant variation between episodes. Due to the fact that an episode spans the course of a year, there are seasonal changes in the weather features, electricity use on weekdays is distinct from weekends, and both solar irradiance and temperatures vary periodically by hour. For these reasons, random separation is unlikely to generate two representative distributions. Comparing consecutive months, weeks, or days often results in excessively large MMDs with low probability values, and so did splitting the data into even and odd time steps (hours). The MMD test indicated that the samples were from two distinct distributions, with $p$-values below 0.05. This means that the previous methods of splitting the data did not produce representative distributions. These separation methods biased the data. Instead, randomly splitting the samples for each day equally produced two representative sets, which MMD indicates are of the same distribution with large $p$ values. This entails assigning 12 random indices from each day to one distribution, and those remaining to another distribution. This ensures that each distribution has an identical number of samples from each day, and randomly sampling each day reduced bias in the selection. Repeating this for 100 pairs of distributions provides a distribution of MMDs and probability values $p$ to compare clean and adversarial observations. For an attack to pass the test, the MMD must be within or below this baseline distribution of MMDs, and within the distribution of probability values. For the latter, we can calculate the percentile of a probability value in the baseline distribution. 

The stealthy PGD attack exhibited a non-negligible  power consumption increase of nearly 10\%, while the MMD is lower than any baseline value and the p-value is high, the SAC is not robust to stealthy attacks. However, the adversarial regret decreased significantly when the adversarial budget was restricted. This effect is even more pronounced in the results of the discrete PPO, as the only attack which both have an MMD lower than the baseline and high $p$-values only increased power consumption by 3\%. These results show that the discrete PPO is robust to stealthy attacks.

Of the attacks which pass the MMD, only the stealthy PGD attack had both a significant adversarial regret and had mean absolute feature variations less than twice the original.  The analysis so far suggests that the stealthy PGD attack is unlikely to be detected. But comparing the adversarial and original \textbf{absolute inter-observation feature variations}, the MMD test assigned a value of $3.5\times10^{-3}$ and a probability value of 0\%. This suggests that attacks with such limited budgets produced convincing adversarial observations but could be identified by comparing them to earlier observations in a time series. SotA adversarial attacks are not designed to be consistent between samples. However, there are caveats:
\begin{itemize}
    \item The MMD test was run under perfect conditions, where the original data was known. A defender could only compare the current distribution with historical data, and the current and past distributions can change for benign reasons. For example, changing power consumption habits from the proliferation of electric vehicles and shifting local weather patterns.
    \item The MMD test only compares two distributions, and this distribution for high variation adversarial features overlap with the originals. This makes identifying individual or small set of adversarial observations a promising subject for future research. Furthermore, a LAA may not require many altered observations to cause a power consumption spike that affects the grid. Because CityLearn does not model loads on the wider power grid, exploring the number of adversarial observations required for such an LAA is out of scope. Thus, the defender must detect a small number of adversarial observations for detection to be a viable defence.
\end{itemize}
These results show that adversarial observations can be detected in aggregate from a time series. There are clear statistical differences in the absolute feature variations between adversarial observations. Exploiting this for an effective defence requires detecting a small number of individual observations and a high-confidence model of how clean observations should behave. High confidence is critical because a detection system with a large false-positive rate risks being ignored. As a white box attack requires a well resourced adversary, the defender must be similarly well resourced to detect the adversarial perturbations. 

\subsubsection{Summary}

Stealth is a significant advantage of adversarial examples, but evading detection significantly constrains the adversarial budget. Limits on budget similarly limit adversarial regret. This section shows that a well constrained attack produces adversarial observations which are statistically indistinguishable from the originals, according to the MMD test. The perturbation boundary $\epsilon$ can be selected for each feature such that: observations appear plausible, the mean variation between observations does not significantly increase, and the maximum mean absolute variation does not exceed the original maximum value. Detecting such an attack is non-trivial, but MMD test results on the distribution of the absolute differences between sequential observations do not exclude the possibility of detection.

\subsection{Robust Training}

\subsubsection{Robust action space}
Training a continuous PPO proved far more difficult than either the discrete PPO and SAC. The continuous PPO tended to get stuck in a local minima where it learned never to fully charge than discharge the battery, so it had no effect on the environment. Achieving a similar electricity consumption KPI required 80 times as many training episodes and a different reward function. The solar penalty reward penalizes the agent for storing energy when consumption is high \cite{nweye2023citylearn}. 

\begin{table}
    \centering
\caption{Adversarial Regret of Stealthy Bifurcated PGD Attack for Different Agents. These results show that regardless of the RL algorithm, a continuous action space is less robust than discrete.}
\label{tab:tri-agent-regret}
    \begin{tabular}{|c|c|c|c|}\hline
 \multicolumn{4}{|c|}{Adversarial Regret of Stealthy Bifurcated PGD Attack for Different Agents}\\\hline \hline 
         KPI&  Discrete PPO&  Continuous PPO& SAC\\ \hline 
         Electricy Consumption 
&  0.025&  0.104& 0.085\\ \hline 
         
Daily Peaks&  0.06&  0.093& 0.273\\ \hline 
         Ramping&  0.142&  0.156& 0.123\\ \hline
    \end{tabular}

\end{table}
Table \ref{tab:tri-agent-regret} demonstrates that the action space makes the discrete PPO robust to adversarial attacks, as the continuous PPO and SAC exhibit significantly higher adversarial regrets. Furthermore, the attack's effect on the discrete PPO's power consumption was relatively small. This is particularly useful for defenders because by testing for the optimal number of bins, the discrete PPO's performance was nearly identical to the SAC for identical training times. There was no compromise in performance for the robustness gained from the discrete action space. 

\subsubsection{ATLA}
\begin{figure}
    \centering
    \includegraphics[width=1\linewidth]{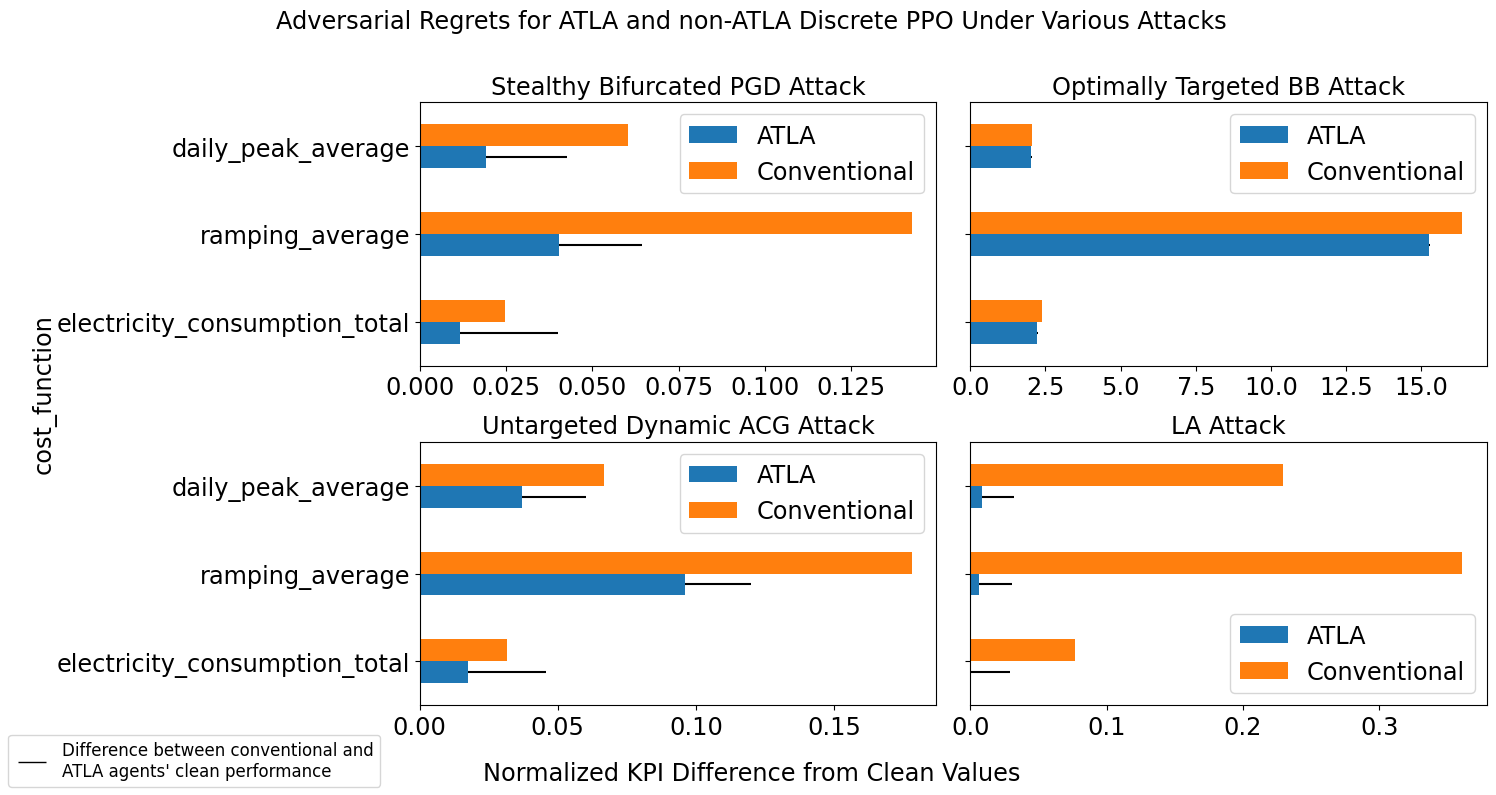}
    \caption{Histogram comparing the adversarial regrets of agents trained conventionally and with ATLA under various attacks. The black bar represents the difference in clean performance between the ATLA and non-ATLA agent, which indicates instances where the regret is smaller for the ATLA agent, but its reduced clean performance means the non-ATLA agent still performs better for that KPI. While the ATLA agent's adversarial regret is smallest in all cases, it still consumes more energy than the conventionally trained agent when attacked with untargeted adversarial examples. The stealthy attack is the bifurcated PGD attack, with $\epsilon=0.03$ masked temporal and solar generation features, and scaled $\epsilon$ for net electricity consumption.}
    \label{fig:ATLA-conv-adv-regret}
\end{figure}

The KPIs shown in Figure \ref{fig:ATLA-conv-adv-regret} demonstrate that ATLA nearly eliminated the effects of an LA attack, indicating that ATLA was successful in training an agent with a robust policy using the $B(s)$ in Table \ref{tab:B(s)}. The question of interest was then if the robust policy decreased the adversarial regret for gradient-based attacks, and if ATLA is a viable defense against such attacks. Figure \ref{fig:ATLA-conv-adv-regret} also shows the adversarial regrets for both the conventional and ATLA agents for various attacks:
\begin{itemize}
    \item The stealthy bifurcated PGD attack will expose the performance of both agents under an attack that is difficult to detect in terms of both feature variations and MMD. 
    \item Optimally targeted BB represents the most powerful, though detectable, attack. 
    \item The dynamic distortion ACG attack is included as a direct (non-bifurcated) attack, which also has a low MMD.
    \item LA attack, which ATLA is specifically designed to resist \cite{Zhang_Chen_Boning_Hsieh_2021}. It outperforms methods which do not attack the neural network function approximator, like those above, and instead exploit weaknesses in the victim agent's policy. 
\end{itemize}
This plot also shows the difference in clean performance between the ATLA and conventional agents, to contextualize the adversarial regrets. While in all cases the ATLA agent loses less performance from attacks, in the case of power consumption it is negated by the ATLA agent's inferior clean performance. Despite this, the ATLA agent performs better in terms of the other KPIs while attacked. 
\begin{table}
    \centering
\caption{The perturbation space for each category of features in ATLA. Features are min-max normalized in [0,1]. This is the absolute amount the adversary can change each feature or the adversarial budget. }
\label{tab:B(s)}
    \begin{tabular}{|c|l|} \hline 
 \multicolumn{2}{|c|}{\textbf{ATLA Adversarial budget $B(s)$ by Feature Category}}\\ \hline 
 Feature&Perturbation\\ \hline 
         Outdoor Dry Bulb Temperature &0.16
\\ \hline 
         Outdoor Relative Humidity&0.36\\ \hline 
         Diffuse Solar Irradiance &0.33\\ \hline 
         Direct Solar Irradiance&0.45\\ \hline 
         Carbon Intensity&0.17\\ \hline 
         Non-Shiftable Load&0.37\\ \hline 
         Solar Generation&$1.5\times 10^{-3}$\\ \hline 
         Electrical Storage SoC &0.32\\ \hline 
         Net Electricity Consumption&$3.3\times 10^{-3}$\\ \hline 
         Electricity Pricing&0.52\\ \hline
    \end{tabular}

\end{table}

These results show that ATLA's usefulness for defending against attacks is nuanced, and depends on the threat environment in which the system operates. Using ATLA on a single agent in CityLearn does reduce the costs, ramping, and peaks during attacks, however these KPIs are larger during normal operation. Based on these results, ATLA is not recommended in general to defend against these white-box gradient-based attacks in CityLearn. 

ATLA does reduce the adversarial regret experienced by a victim agent, but the trade-off in clean performance did not justify the additional robustness for this task. Thus the decision to use ATLA is application specific and should reflect the threat model in which the agent would operate. While ATLA may not be beneficial in a domestic setting, it could be sensible for a hydro-electric or mass energy storage facility who's energy consumption could significantly affect the power grid.

\subsection{Snooping Attack}
\begin{figure}
    \centering
    \includegraphics[width=1\linewidth]{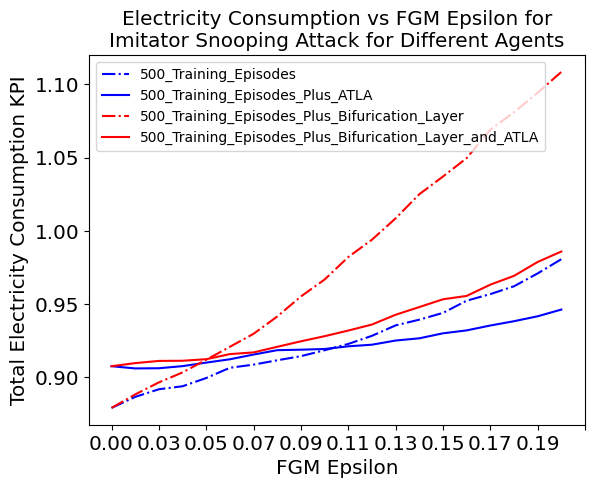}
    \caption{Comparison of electricity consumption KPI for direct and bifurcated FGM  snooping attacks with a range of $\epsilon$, for discrete PPO agents trained conventionally and with the ALTA method. The line symbols indicate the agent, while the colour indicates the attack. This plots shows the trend between adversarial budget and regret. Solid lines represent ATLA training. While the robustness offered by ATLA is insignificant compared to its reduction in clean performance for energy consumption, the corresponding adversarial regret for bifurcated attacks is significantly reduced.}
    \label{fig:ATLA-cost-energy-vs-snooping-fgsm}
\end{figure}

Snooping attacks were conducted using the FGM, as per the methodology of \cite{Inkawhich_Chen_Li_2020}. Because FGM is a weaker attack compared to iterative methods, it requires a larger $\epsilon$ for a similar adversarial regret. The relationship between the adversarial budget and regrets  for the snooping attack on the discrete PPO are shown in Figure \ref{fig:ATLA-cost-energy-vs-snooping-fgsm}. When random noise is compared to a snooping attack with the same $\epsilon$, the random attack requires almost 10 times the adversarial budget for a similar ASR. Random noise did not have a significant adversarial regret for any perturbation size tested, up to $\epsilon = 0.2$. With $\epsilon = 0.05$ the power consumption for bifurcated FGM is comparable to the white-box ACG dynamic distortion attack. The discrete PPO is robust in that it requires $\epsilon \approx 0.13$ before power consumption is equivalent to no controller, and is increased by over 10\%. 
By comparing the slope of the direct and bifurcated attacks, the impact of the former becomes apparent. The bifurcation method significantly increases the cost and power consumption for the conventionally trained agent, though the ATLA trained agent sees a much smaller difference. 
Unlike the white-box attacks of the previous section, ATLA provides a significant resistance to the most powerful technique used in this section. The bifurcated snooping attack produced a significantly higher adversarial regret for the conventional agent compared to the ATLA agent. This is the same to a smaller extent for the direct snooping attack, and this was partially counteracted by the ATLA agent's lower clean performance. Because black box attacks require fewer prerequisites for the attacker, they are more likely to be encountered. These attacks are also easier to detect, as previous works have shown that adversarial samples crafted with FGM are detectable through statistical techniques \cite{Grosse_Manoharan_Papernot_Backes_McDaniel_2017}. Furthermore, this work found that attacks with $\epsilon > 0.03$ are statistically detectable in CityLearn, while snooping attacks require four times that to undo the benefit of the DRL controller for a discrete agent. 

\begin{figure}
    \centering
    \includegraphics[width=1\linewidth]{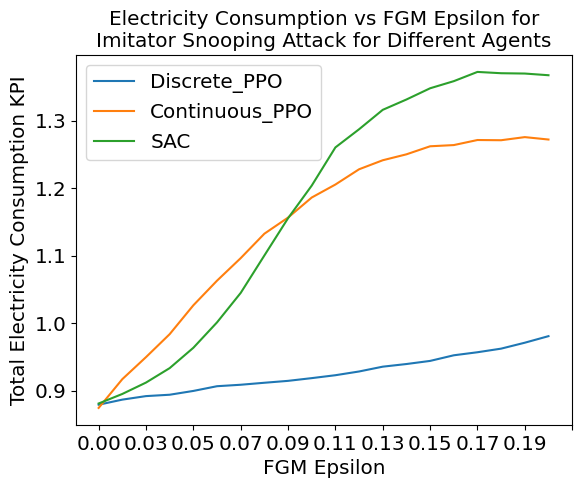}
    \caption{Comparison electricity consumption KPI for bifurcated FGM  snooping attacks with a range of $\epsilon$, for discrete and continuous PPO, and SAC agents.  This figure compares the trend of adversarial budget and regret between various DRL algorithms and action spaces. This figure demonstrates that the discrete PPO is significantly more robust than either agent with a continuous action space, even without ATLA training.}
    \label{fig:discrete-cont-snooping}
\end{figure}

The performance of the discrete and continuous  PPO, and SAC under a bifurcated snooping attack are compared in Figure \ref{fig:discrete-cont-snooping}. While $\epsilon \approx 0.13$ is required to remove the benefit of the discrete PPO in terms of power consumption, the same happens for the SAC for $\epsilon \approx 0.06$ and $\epsilon \approx 0.05$ for the discrete PPO. At this smaller value of $\epsilon$ the adversarial regret for the discrete PPO is only 0.03. Even without ATLA training, the discrete action space significantly improves the robustness of the PPO to the black-box  snooping attack.

An adversary can achieve a significant reduction in performance, while only observing how the victim behaves in their environment and exploiting vulnerable sensor systems. Unlike the finding in \cite{Zhang_Chen_Boning_Hsieh_2021}  where the LA outperformed the snooping attack, combining the snooping attack with the bifurcation method outperforms the LA attack while using a smaller adversarial budget. The snooping attack is gradient-based like a white box attack, which typically outperform black box attacks when a gradient is available. The bifurcation method enhances the loss function for gradient-based attacks, improving the ratio of adversarial regret to budget. The bifurcated snooping attack had a power consumption comparable to the LA attack with $\epsilon=0.07$, which is much smaller than the $B(s)$ for the LA attack, for all except solar generation and net electricity consumption features.  Designers should invest in detection systems to counter such methods, and ATLA could be useful depending on the trade-off between performance and robustness.

\section{Conclusion}
\label{sec:conclusion}
This work proposes a novel adversarial attack technique for continuous control, the GDL with a bifurcation layer. By implementing our GDL loss, the adversarial regret for SotA attacks was doubled for the continuous control agent in CityLearn. The bifurcation method can be used for continuous and discrete outputs, even for attacks which only support ANN with multiple outputs. This technique was validated using identical attacks without the bifurcation layer for both classification/discrete actions and regression/continuous actions. To validate the bifurcation method for continuous action spaces, this work implemented PGD for a regressor.

We studied the detection of adversarial observations and concluded that in this setting, with a carefully selected budget, distributions of adversarial observations, which are not significantly different from the originals, cause a significant adversarial regret. Conversely, adversarial observations form a distinct distribution when considering the absolute difference between subsequent observations in a time-series. The latter result could be used in the detection of adversarial observations. 

This work found that the choice of off-the-shelf DRL algorithm and action space can significantly affect the agent's robustness. When the adversarial budget was decreased to the point that the perturbations were considered stealthy, the attack had a minimal effect on the discrete PPO but still caused a significant adversarial regret for the SAC and continuous PPO. The effects of adversarial observations, even in the best-case white-box scenario, are significantly reduced using a discrete action space with a PPO. There was not a significant difference in clean performance between the PPO and SAC, meaning there was no concession for the additional robustness.

Alternating Training with Learned Adversary (ATLA) is an effective defence against strong black box attacks targeting the ANN function approximator. Combining the bifurcation method with the snooping attack significantly increased the adversarial regret for the conventionally trained PPO, but the increase was less than half for the PPO trained using ATLA.


\bibliographystyle{IEEEtran}
\bibliography{bibliography}



 




\vfill

\end{document}